\documentclass[sigconf,letter,authorversion,nonacm]{acmart}

\usepackage{geometry}
\usepackage{listings}
\usepackage{multicol} 
\usepackage{float}
\usepackage{algorithm}
\usepackage{algpseudocode}
\AtBeginDocument{%
  \providecommand\BibTeX{{%
    \normalfont B\kern-0.5em{\scshape i\kern-0.25em b}\kern-0.8em\TeX}}}

\copyrightyear{2023}
\acmYear{2023}
\setcopyright{acmlicensed}\acmConference[GeoAI '23]{6th ACM SIGSPATIAL International Workshop on AI for Geographic Knowledge Discovery}{November 13, 2023}{Hamburg, Germany}
\acmBooktitle{6th ACM SIGSPATIAL International Workshop on AI for Geographic Knowledge Discovery (GeoAI '23), November 13, 2023, Hamburg, Germany}
\acmPrice{15.00}
\acmDOI{10.1145/3615886.3627742}
\acmISBN{979-8-4007-0348-5/23/11}

\begin{document}

\title[FLEE-GNN]{FLEE-GNN: A Federated Learning System for Edge-Enhanced Graph Neural Network in Analyzing Geospatial Resilience of Multicommodity Food Flows}

\author{Yuxiao Qu$^{1,2}$, Jinmeng Rao$^{2}$, Song Gao$^{2}$, Qianheng Zhang$^{2}$, Wei-Lun Chao$^{3}$, Yu Su$^{3}$, Michelle Miller$^{4}$, Alfonso Morales$^{5}$, Patrick R. Huber$^{6}$}
\affiliation{%
  \institution{ 
  $^1$Department of Computer Science, Carnegie Mellon University\\
  $^2$Geospatial Data Science lab, University of Wisconsin, Madison\\ $^3$Department of Computer Science and Engineering, Ohio State University\\
  $^4$Center for Integrated Agricultural Systems, University of Wisconsin, Madison\\ 
  $^5$Department of Planning and Landscape Architecture, University of Wisconsin, Madison\\
  $^6$  Institute of the Environment, University of California, Davis\\
  }
  \country{ }
  }
\email{Email: song.gao@wisc.edu}

\renewcommand{\shortauthors}{Qu et al.}

\begin{abstract}
Understanding and measuring the resilience of food supply networks is a global imperative to tackle increasing food insecurity. However, the complexity of these networks, with their multidimensional interactions and decisions, presents significant challenges. This paper proposes FLEE-GNN, a novel Federated Learning System for Edge-Enhanced Graph Neural Network, designed to overcome these challenges and enhance the analysis of geospatial resilience of multicommodity food flow network, which is one type of spatial networks. FLEE-GNN addresses the limitations of current methodologies, such as entropy-based methods, in terms of generalizability, scalability, and data privacy. It combines the robustness and adaptability of graph neural networks with the privacy-conscious and decentralized aspects of federated learning on food supply network resilience analysis across geographical regions. This paper also discusses FLEE-GNN's innovative data generation techniques, experimental designs, and future directions for improvement. The results show the advancements of this approach to quantifying the resilience of multicommodity food flow networks, contributing to efforts towards ensuring global food security using AI methods. The developed FLEE-GNN has the potential to be applied in other spatial networks with spatially heterogeneous sub-network distributions.
\end{abstract}



\maketitle

\section{Introduction}

According to the State of Food Security and Nutrition in the World 2023 report jointly prepared by FAO, IFAD, UNICEF, WFP, and WHO~\cite{unicef2023state}, the world is moving backwards in its efforts to end hunger, food insecurity, and malnutrition in all its forms. The number of people experiencing acute food insecurity and requiring urgent food, nutrition, and livelihood assistance increased for the fourth consecutive year in 2022. 
The need to transform agrifood systems for increased resilience is urgent, as it can help provide nutritious foods at lower costs and ensure affordable, healthy diets for everyone in a sustainable and inclusive manner. Within the agrifood systems, food supply networks are pivotal in upholding global food security and facilitating the transit, dissemination, and sale of food. It's imperative that these networks demonstrate resilience and sturdiness~\citep{suweis2015resilience,miller2021identifying,karakoc2021complex}. 

However, the complexity inherent in them, arising from diverse food needs, shipment timeframes and costs, promotional strategies, cultural and environmental considerations, among others, complicates the assessment of their durability and adaptability~\cite{chaturvedi2014securing, yadav2022systematic}. Given the intricate nature of food supply networks, the concept of resilience is often interpreted in diverse ways by different individuals and groups~\cite{fair2017dynamics,puma2019resilience,tu2019impact,karakoc2021complex}. The term "resilience" in this study predominantly pertains to the capacity of the food flow networks to sustain essential food supplies across geographical regions despite potential disruptions. In general terms, a node within the food flow network is deemed less resilient if its dependence is heavily skewed towards a sole supplier, a far-off supplier, a singular consumer, or one specific commodity \cite{rao2022measuring,inman2014product}.

Currently, resilience measurement methods such as the network topological characteristics and motifs \cite{dey2019network,gao2016universal,karakoc2021complex}, geospatial knowledge graph (GeoKG) and entropy-based metrics \cite{rao2022measuring} have been employed to clarify the semantics of spatial networks~\cite{barthelemy2011spatial} (e.g., a multicommodity flow network between regions) and to evaluate its resilience. However, these mostly centralized methods encounter some limitations:
\begin{itemize}
	\item Robustness: the existing model is vulnerable to missing data or poorly curated data.
	\item Requirement for centralized data access: the conventional entropy-based method necessitates access to the entire dataset, which is problematic given the recent emphasis on privacy concerns.
\end{itemize}

As a decentralized approach, Federated Learning (FL) has been adopted in various agricultural applications and supply chain systems. Existing efforts include utilizing federated learning to facilitate agricultural data sharing~\cite{durrant2022role}, combat food fraud~\cite{gavai2023applying}, and predict supply chain risks~\cite{zheng2023federated}, yielding promising results. Hence, to address the issues exhibited in centralized methods, we propose a Federated Learning System for Edge-enhanced Graph Neural Network (FLEE-GNN). This innovative approach aims to improve the robustness and decentralize the measurements related to the geospatial resilience of multicommodity food flows, which refers to the capacity of a geographical region to withstand and recover from a variety of environmental, social, and economic challenges and disruptions on food supplies. By fusing the generalization capabilities of Graph Neural Networks (GNNs) \citep{scarselli2008graph, mpgnn, zhang2021magnet} with the decentralized features of federated learning \cite{chen2020fedbe}, the FLEE-GNN offers a more resilient, flexible, and privacy-aware solution than centralized approaches for analyzing multicommodity flow networks and beyond.

This paper makes the following key contributions:
\begin{itemize}
	\item We designed an edge-enhanced graph neural network and a federated learning framework to measure network resilience across geographical regions in a decentralized manner.
	\item We introduced an adjustable data generator to overcome data scarcity in measuring food supply resilience with deep learning methods.
	\item We conducted a comprehensive analysis of the performance of both centralized and decentralized machine learning methods, comparing them to an entropy-based approach using the U.S. food flow data.
\end{itemize}

The rest of the paper is structured as follows:
\begin{itemize}
	\item Section 2 presents the elements of FLEE-GNN, encompassing GNN architecture, the federated learning framework, and a data generator that aids in addressing data shortages in the realm of multicommodity food flows.
	\item Section 3 outlines the datasets and models we assessed in the experiments.
	\item Section 4 delves into the results of prediction error, rank evaluation, and robustness across various models and datasets.
	\item Section 5 outlines the directions of our future research, spotlighting possibilities like unsupervised learning and enhanced GeoKG embedding.
	\item Section 6 provides a conclusion drawn from this  study.
\end{itemize}

\section{Methods}

\subsection{System Design}

In this section, we introduce the components of FLEE-GNN, which include a message-passing-based GNN and a specially designed federated learning system to handle cases where there is no freight transportation or information exchange between geographical regions (``hypothesized disruptions'') in the U.S.

\subsubsection{Edge-enhanced Graph Neural Network}

\begin{figure}[h]
	\centering
	\includegraphics[width=8cm]{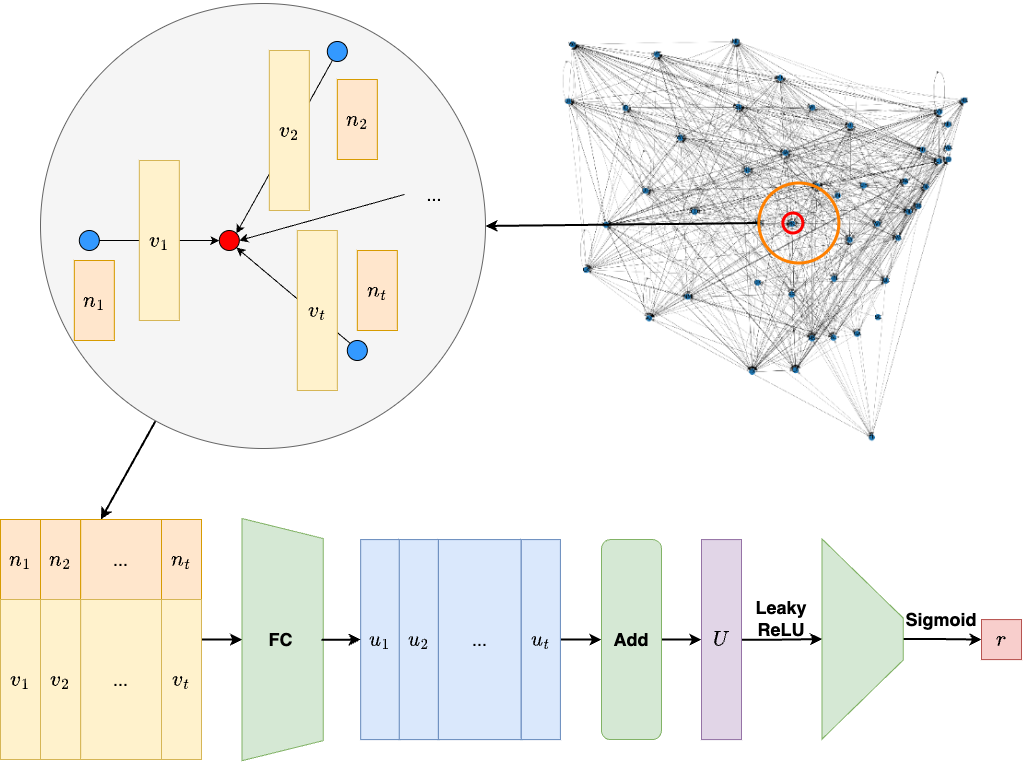}
	\caption{The architecture for the edge-enhanced graph neural network}
	\label{fig:architecture}
\end{figure}

As shown in Figure \ref{fig:architecture}, We have adapted a message-passing model to develop the edge-enhanced GNN. This model accepts a graph, $G$, as its input. Here, $G = (V, E)$ where $V$ represents the set of all nodes in a graph. For the purpose of this work, we consider each U.S. state as a node in the spatial network. $E$ is the set of all commodity food transported from a source state, $s$, to a destination state, $d$. Each transport of a commodity type between a pair of origin and destination is characterized by its monetary value $v$, weight tonnage $t$, and average transportation distance in mile $a$. It denotes as $E= \{(s, d, v_s, t_s, a_s)_i | i \in n\}$, where $n$ is the number of edges $|E|$.  

In this study, we use the Standard Classification of Transported Goods (SCTG) Codes from the U.S. Commodity Flow Survey (CFS) to categorize each food commodity type:
\begin{itemize}
	\item 01: Animals and fish (live)
	\item 02: Cereal grains (includes seed)
	\item 03: Agricultural products (excluding animal feed, cereal grains, and forage products)
	\item 04: Animal feed, eggs, honey, and other products of animal origin
	\item 05: Meat, poultry, fish, seafood, and their preparations
	\item 06: Milled grain products, preparations, and bakery products
	\item 07: Other prepared foodstuffs, and fats and oils (CFS10)
	\item 08: Alcoholic beverages and denatured alcohol (CFS20)
\end{itemize}

For each node, we relay its neighboring node features, which include the location information (i.e., latitude and longitude), as well as the edge features, which encompasses the value, tonnage, and average transportation miles of all commodity food types transported from a source state to a destination state. The features of the destination state itself are captured by an edge linking from itself. If there is no transportation within the state, this results in zero edge feature values. Conversely, if transportation has occurred within the state, it results in non-zero edge feature values.

For illustration, consider the value of agricultural products (SCTG 03) from Alabama (AL) to Georgia (GA), which stands at 145 dollars per ton. The value for other prepared foodstuffs (SCTG 07) from AL to GA is 1497 dollars per ton. In 2012, 197 tons of agricultural products were transported from AL to GA, with an average transportation distance of 249 miles. Conversely, 613 tons of other prepared foodstuffs were transported from AL to GA, averaging a distance of 152 miles. These are the only two food types transported from AL to GA. Assuming AL is the $i$'th neighbor for GA, then the edge features $v_i$ during computation are represented as a vector with 24 dimensions, with specific dimensions as $\{V_3: 145, T_3: 197, A_3: 249, V_7: 1497, T_7: 613, A_7: 152\}$. The remaining dimensions are zero:
\begin{table}[h]
	\begin{tabular}{c|c|c|c|c|c|c|c|c}
		... & $\textbf{V}_3$ & $\textbf{T}_3$ & $\textbf{A}_3$ & ... & $\textbf{V}_7$ & $\textbf{T}_7$ & $\textbf{A}_7$ & ... \\ \hline
		0   & 145        & 197        & 249        & 0   & 1497       & 613        & 152        & 0  
	\end{tabular}
\end{table}

Every message, denoted as $[n_i, v_i]$ (a combination of edge and node features), passes through a fully connected layer in the neural network (see Figure \ref{fig:architecture}). It then merges into a latent vector, $u_i$, encapsulating the information within the transitions from the source to destination states. This condensation into a lower dimension bolsters computational efficiency. After procuring all latent vectors for the destination state's neighbors, every message undergoes aggregation via an summation function. This is then transmuted into another latent vector, $U$, signifying all data relevant to import resilience for the destination state. This vector is then projected into a scalar value, which undergoes another fully connected layer, followed by a sigmoid activation function, ensuring an output range between 0 and 1, which represents the predicted resilience score.

This design ensures that both node and edge information is leveraged at each processing stage, aiming to provide a richer representation for various graph-based tasks.

\subsubsection{Federated Learning}
\begin{figure*}[h]
	\centering
	\includegraphics[width=\textwidth]{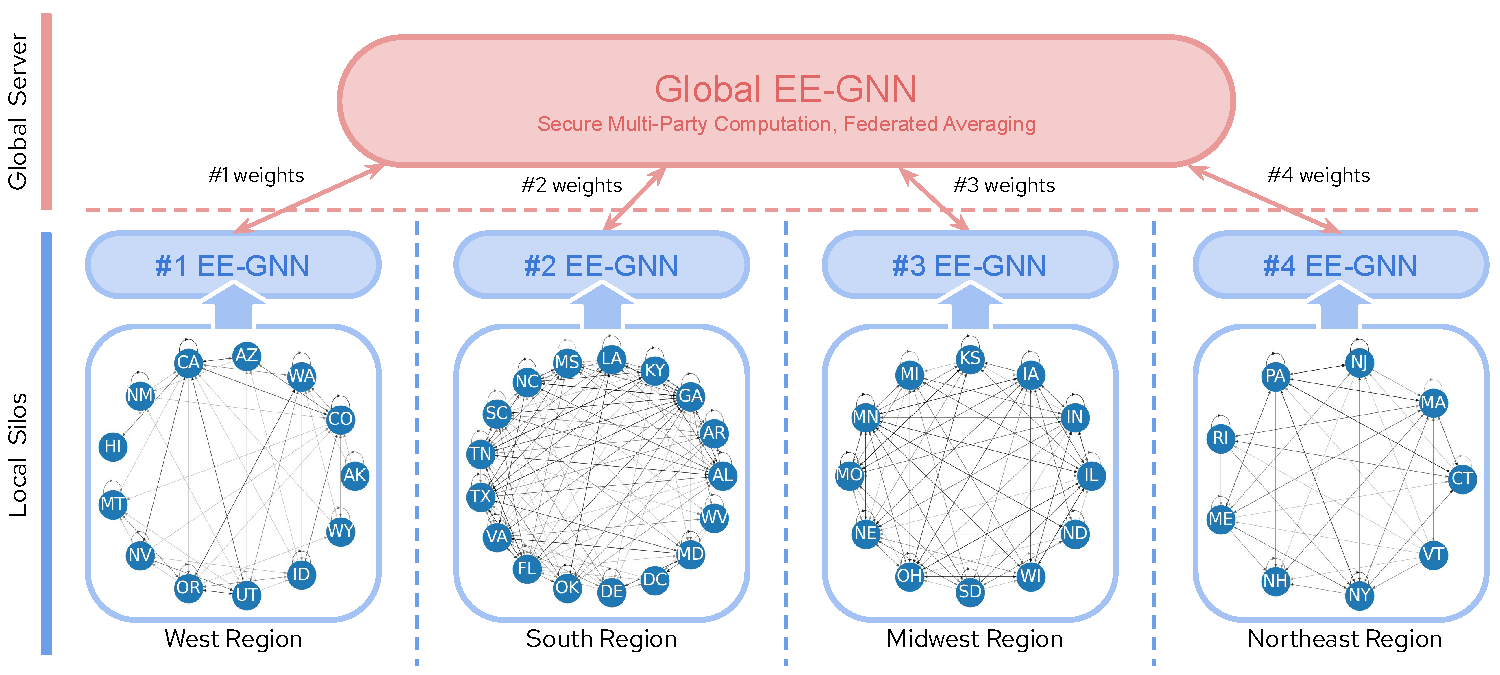}
	\caption{The federated learning architecture for the edge-enhanced graph neural networks}
	\label{fig:fl}
\end{figure*}

This study introduces a cutting-edge federated learning framework to assess the geospatial resilience of multicommodity food flow as shown in Figure \ref{fig:fl}. 

The federated learning approach embodies the concept of decentralized training \cite{Konecn2016FederatedOD}. Here's a detailed overview of the process:

\begin{enumerate}
	\item Initialization: The central server initializes a global model. This model is then dispatched to all distributed region nodes (or 'silos') participating in the federated learning network. In this research, each silo refers to each geographical region in the U.S. and there are four regions (i.e., West, South, Midwest, and Northeast) in our experiments. 
	\item Local Training: Each node computes its model locally using its own dataset. Importantly, no raw data is exchanged or shared between nodes or with the central server. Training is performed over a pre-defined number of epochs (e.g., 100 epochs) to adjust and refine the model using local data.
	\item Local Model Update: At the end of local training, each node computes a summary of its model updates. These are essentially changes or differences compared to the global model.
	\item Aggregation at Central Server: Nodes send their model updates to the central server periodically (e.g., after every 10 epochs). The central server then aggregates these updates using methods like weighted averaging.
	\item Global Model Refinement: After aggregation, the central server refines the global model parameters, which benefit from the knowledge gathered from all distributed nodes.
	\item Dispatching Updated Model: The improved global model is then sent back to all regional nodes. This cyclic process of local training, aggregation, and global update continues until the model converges or another pre-defined criterion to stop is met.
\end{enumerate}

\subsection{Data Generation}

The U.S. Bureau of Transportation (BTS) and the U.S. Census Bureau publish the detailed commodity food flow survey (CFS) \footnote{\url{https://www.bts.gov/product/commodity-flow-survey}} once every five years, which presents a data scarcity challenge for the idea of measuring food supply resilience using deep learning methods. To overcome this challenge, we introduce an adjustable data generator (Algorithm \ref{algo:graph_generator}), in conjunction with the previous entropy-based method \cite{rao2022measuring}, to create nearly authentic food flow data. This data generator requires a real commodity flow dataset as input, accompanied by a noise ratio. The noise ratio is defined as the proportion of data we wish to modify based on the actual dataset. We have also delineated three methods to adjust the data:

\begin{algorithm}
	\caption{GRAPH\_GENERATOR}
	\begin{algorithmic}[1]
		\Require 
		$G$ is the true commodity food flow graph with $n$ edges, where $G = (V, E)$ and $n = |E|$, $r$ is the noise ratio
		\Ensure $G$ is the generated commodity food flow graph
		
		\State $n' \gets \left\lfloor \frac{r \times n}{3} \right\rfloor$ \Comment{each operation [REMOVE, CHANGE, ADD] contribute a third of the total noise}
		
		\For{$i = 1$ to $n'$}
		\State $G \gets \text{REMOVE}(G)$
		\State $G \gets \text{CHANGE}(G)$
		\State $G \gets \text{ADD}(G)$
		\EndFor
		
	\end{algorithmic}
	\label{algo:graph_generator}
\end{algorithm}

\begin{enumerate}
	\item ADD (Algorithm \ref{algo:add}): Introduce a transportation event for food type $c'$ from source $s'$ to destination $d'$ with a random value $v'$, tonnage $t'$, and average transportation miles $a'$ sampled from its corresponding distribution in the original dataset. If the sampled tuple $(s', d', c')$ is already present, we sample another one.
	\item REMOVE (Algorithm \ref{algo:remove}): Randomly select an edge and expunge it from the current food flow.
	\item CHANGE (Algorithm \ref{algo:change}): Randomly select an edge and resample value $v'$, tonnage $t'$, and average transportation miles $a'$ from its corresponding distribution in the original dataset.
\end{enumerate}

\begin{algorithm}
	\caption{ADD}
	\begin{algorithmic}[1]
		\Require 
		$G$ is a directed graph with $n$ edges, where $n = |E|$, $G = (V, E)$, and $E = \{(s_i, d_i, c_i, v_i, t_i, a_i)|i \in [1, n]\}$. $C$ is a set of commodity food
		\Ensure $G'$ is a directed graph with $(n+1)$ edges
		\State $MAX_V \gets \max_{i=1}^{n} v_i$, $MIN_V \gets \min_{i=1}^{n} v_i$
		\State $MAX_T \gets \max_{i=1}^{n} t_i$, $MIN_T \gets \min_{i=1}^{n} t_i$
		\State $MAX_A \gets \max_{i=1}^{n} a_i$, $MIN_A \gets \min_{i=1}^{n} a_i$
		\Repeat
		\State $s' \gets \text{sample a source node from } V$
		\State $d' \gets \text{sample a destination node from } V$
		\State $c' \gets \text{sample a commodity food from } C$
		\Until{there is no edge in $G$ that goes from $s'$ to $d'$ with $c = c'$}
		\State $v' \gets \text{sample a real number from } [MIN_V, MAX_V]$
		\State $t' \gets \text{sample a real number from } [MIN_T, MAX_T]$
		\State $a' \gets \text{sample a real number from } [MIN_A, MAX_A]$
		\State $E' \gets E \cup \{(s', d', c', v', t', a')\}$
		\State $G' \gets (V, E')$
	\end{algorithmic}
	\label{algo:add}
\end{algorithm}

\begin{algorithm}
	\caption{REMOVE}
	\begin{algorithmic}[1]
		\Require $G$ is a directed graph with $n$ edges, where $n=|E|$, and $G=(V, E)$
		\Ensure $G'$ is a directed graph with $(n-1)$ edges
		\State $i \gets \text{sample an integer from } 1 \text{ to } n$
		\State $E' \gets E \setminus \{(s_i, d_i, c_i, v_i, t_i, a_i)\}$
		\State $G' \gets (V, E')$
	\end{algorithmic}
	\label{algo:remove}
\end{algorithm}

\begin{algorithm}
	\caption{CHANGE}
	\begin{algorithmic}[1]
		\Require 
		$G$ is a directed graph with $n$ edges, where $n = |E|$, and $G = (V, E)$
		\Ensure $G'$ is a directed graph with $n$ edges
		\State $MAX_V \gets \max_{i=1}^{n} v_i$, $MIN_V \gets \min_{i=1}^{n} v_i$
		\State $MAX_T \gets \max_{i=1}^{n} t_i$, $MIN_T \gets \min_{i=1}^{n} t_i$
		\State $MAX_A \gets \max_{i=1}^{n} a_i$, $MIN_A \gets \min_{i=1}^{n} a_i$
		\State $I \gets \text{sample an integer from } 1 \text{ to } n$
		\State $(s', d', c', v', t', a') \gets (s_i, d_i, c_i, v_i, t_i, a_i)$
		\State $v' \gets \text{sample a real number from } [MIN_V, MAX_V]$
		\State $t' \gets \text{sample a real number from } [MIN_T, MAX_T]$
		\State $a' \gets \text{sample a real number from } [MIN_A, MAX_A]$
		\State $E' \gets E \setminus \{(s_i, d_i, c_i, v_i, t_i, a_i)\}$
		\State $E' \gets E' \cup \{(s', d', c', v', t', a')\}$
		\State $G' \gets (V, E')$
	\end{algorithmic}
	\label{algo:change}
\end{algorithm}

Given a noise ratio, we evenly apply the three aforementioned operations to generate a new food flow dataset for data augmentation purpose. For instance, if the original datasets contain 100 edges and we have a noise ratio of 0.3, we will add 10 edges, remove 10 edges, and change 10 edges. This results in a dataset with 100 edges, albeit different from the original set.

\section{Experiments}

\subsection{Datasets}

In our experiments, we utilize the above-mentioned U.S. CFS data. The CFS provides
comprehensive information of domestic freight shipments including commodity type, value, weight, distance shipped, origin and
destination, etc. from national-level to state-level. As an example, we focus on the agricultural multi-commodity flows in 2012 and
2017 at differential geographical regions by extracting the data
with SCTG code from 01 to 08. We further process the raw data into two primary datasets:

\begin{enumerate}
	\item Silo Dataset: Assumes that each silo (i.e., geographic region) can only access the links and transitions within its domain.
	
	\item Centralized Dataset: Encompasses all links and transitions, both within and across silos, thus serving as a comprehensive graph of transitions.
\end{enumerate}

The basic graph statistics of the whole food supply network and different sub-networks by regions are illustrated in Table \ref{tab:graph_stats_2012} and Table \ref{tab:graph_stats_2017}.
The sub-networks of food flows in different geographical regions have distinctive graph statistics (e.g., degree, centrality, connectivity), which reflects spatial heterogeneity under different spatial configurations and poses challenges on geospatial resilience modeling and evaluation.

\begin{table*}[h]
	\scalebox{0.9}{
		\begin{tabular}{c|c|c|c|c|c|c}
			\textbf{Area}                              & \textbf{Whole} & \textbf{Whole w/o cross-silo edges} & \textbf{West Silo} & \textbf{Midwest Silo} & \textbf{South Silo} & \textbf{Northeast Silo} \\ \hline
			\textbf{average degree}                    & 63.7255        & 27.7647                             & 16.0000            & 37.1667               & 32.7059             & 22.8889                 \\ \hline
			\textbf{average weighted degrees (VAL)}    & 25797.3333     & 20902.3529                          & 16193.8462         & 28875.5000            & 18695.4118          & 21241.3333              \\ \hline
			\textbf{average degree centrality}         & 1.2745         & 0.5553                              & 1.3333             & 3.3788                & 2.0441              & 2.8611                  \\ \hline
			\textbf{average closeness centrality}      & 0.6153         & 0.1681                              & 0.5533             & 0.8211                & 0.6620              & 0.7492                  \\ \hline
			\textbf{average betweenness centrality}    & 0.0134         & 0.0025                              & 0.0594             & 0.0250                & 0.0355              & 0.0536                  \\ \hline
			\textbf{average node connectivity}         & 13.6184        & 1.2894                              & 2.5897             & 6.9697                & 6.2169              & 3.7917                  \\ \hline
			\textbf{edge connectivity}                 & 1              & 0                                   & 0                  & 3                     & 1                   & 1                      
		\end{tabular}
	}
	\caption{Graph Statistics for 2012 CFS data}
	\label{tab:graph_stats_2012}
\end{table*}

\begin{table*}[h]
	\scalebox{0.9}{
		\begin{tabular}{c|c|c|c|c|c|c}
			\textbf{Area}                              & \textbf{Whole} & \textbf{Whole w/o cross-silo edges} & \textbf{West Silo} & \textbf{Midwest Silo} & \textbf{South Silo} & \textbf{Northeast Silo} \\ \hline
			\textbf{average degree}                    & 49.7647        & 21.0980                             & 13.3846            & 28.8333               & 25.7647             & 13.1111                 \\ \hline
			\textbf{average weighted degrees (VAL)}    & 16365.5294     & 12778.3137                          & 9137.0769          & 17025.3333            & 13682.1176          & 10668.0000              \\ \hline
			\textbf{average degree centrality}         & 0.9953         & 0.4220                              & 1.1154             & 2.6212                & 1.6103              & 1.6389                  \\ \hline
			\textbf{average closeness centrality}      & 0.5515         & 0.1477                              & 0.4289             & 0.7759                & 0.6264              & 0.5110                  \\ \hline
			\textbf{average betweenness centrality}    & 0.0152         & 0.0027                              & 0.0606             & 0.0303                & 0.0336              & 0.1012                  \\ \hline
			\textbf{average node connectivity}         & 11.2208        & 1.1031                              & 1.9295             & 6.5985                & 5.5662              & 1.7639                  \\ \hline
			\textbf{edge connectivity}                 & 0              & 0                                   & 0                  & 4                     & 0                   & 0                      
		\end{tabular}
	}
	\caption{Graph Statistics for 2017 CFS data}
	\label{tab:graph_stats_2017}
\end{table*}

\subsection{Models}

The baseline model, derived from our previous work \cite{rao2022measuring}, employs an entropy-based method using the centralized food flow dataset. Given its access to all links between geographical regions, we designate this model as our ground truth. 

A node-level resilience $R_i$ that comprehensively measures the single-supplier/customer dependence, single-commodity-type dependence, transport distance, and geographic adjacency of the node $i$ is computed as:

\begin{equation}
R_{i} = 1 - D_{i}\frac{\sum_{A \in Agg}V'_{(i, A)}}{V'_{i}}
\label{eq:R_i}
\end{equation}

where $D_{i}$ is the overall single-commodity-type dependence of node $i$ (measured by the Shannon information entropy); $Agg$ is a set of aggregated commodity types; $V'_{(i, A)}$ is the value of aggregated commodity $A$ reflecting single-commodity-type dependence; $V'_{i}=\sum_{A \in Agg}\sum_{c \in A}\sum_{j}V'_{(i\rightarrow j, c)}$ denotes the total commodity value of node $i$ combined with average transport miles and geography adjacency information. The higher the $R_i$, the less dependent node $i$ is on single or geographically distant/non-adjacent supplier/customer or single commodity type, thus the higher the resilience.

We set the developed FLEE-GNN model against the entropy-based method and a centralized GNN (CT-GNN). We analyzed how each model performed under various noise levels by perturbing the 2012 food flow data at noise ratios of 0.1, 0.3, and 0.5, resulting in three datasets of 500 graphs each. Both CT-GNN and FLEE-GNN models underwent training for 100 epochs. Evaluations were then carried out on real food flow data from 2012 and 2017, respectively.

\section{Results}

\subsection{Relative Resilience Difference}

\begin{figure*}[!h]
	\centering
	\begin{minipage}[t]{0.48\textwidth}
		\centering
		\includegraphics[width=\textwidth]{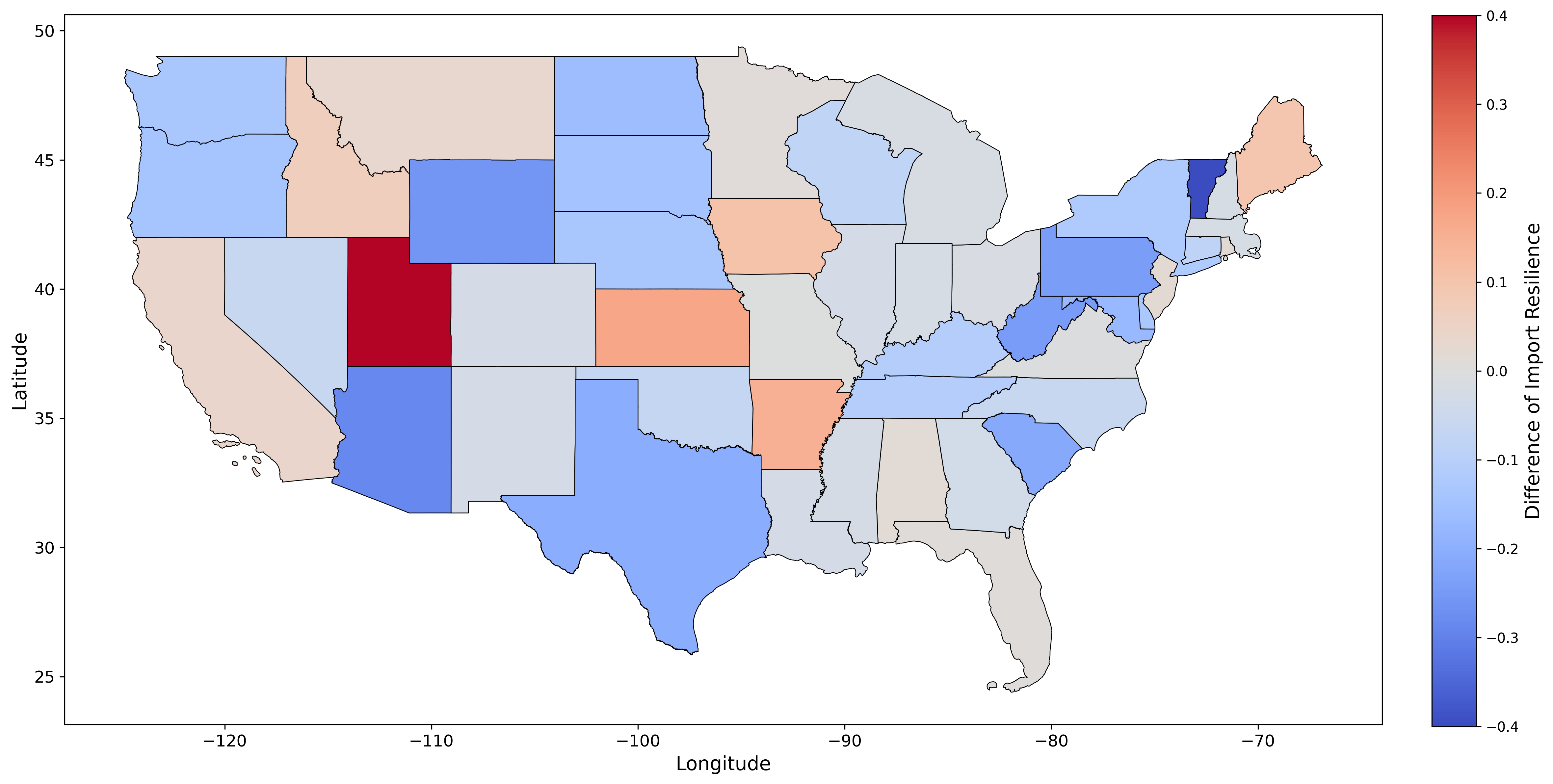}
	\end{minipage}
	\begin{minipage}[t]{0.48\textwidth}
		\centering
		\includegraphics[width=\textwidth]{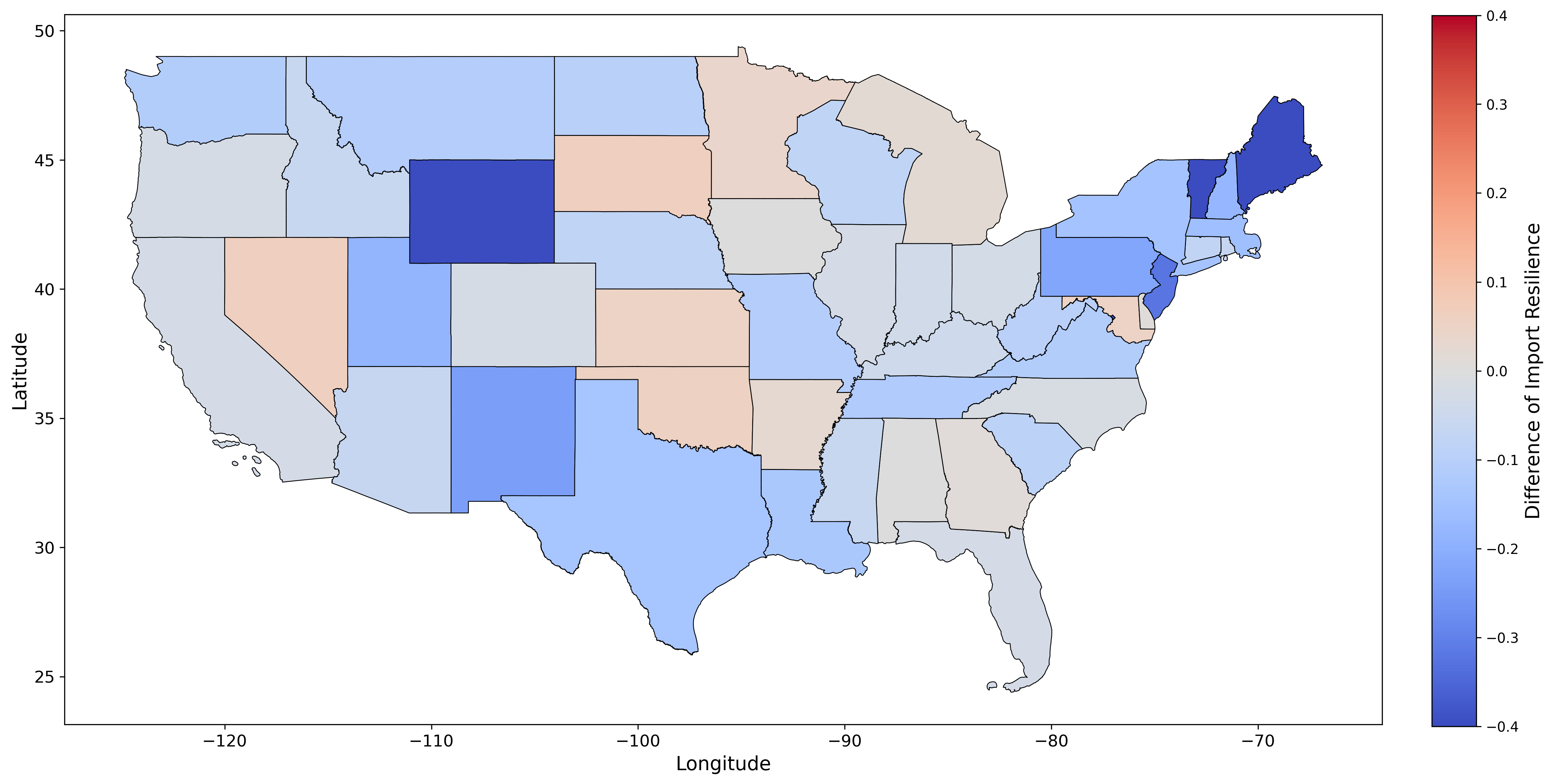}
	\end{minipage}
	\label{entropy}
	\caption{The difference in resilience between the entropy-based method using silo data and the entropy-based method using global data for 2012 (L) and 2017 (R)}
	\label{fig:0.3_entropy}
\end{figure*}

We define the relative resilience difference as the difference between each model output and the entropy-based computation result (ground truth). As seen in Figure \ref{fig:0.3_entropy}, it is evident that the resilience values generated by the entropy-based method within sub-networks only take into account data within a silo, causing the model to underestimate its global resilience in the food supply network. This can be attributed to the omission of transportation between different geographical silos. By neglecting these transportation movements, we effectively remove edges from the original spatial network. As a result, nodes, especially those that lose transportation connections, become more vulnerable to potential disruptive changes imposed on the food systems.

\begin{figure*}[!h]
	\centering
	\begin{minipage}[t]{0.48\textwidth}
		\centering
		\includegraphics[width=\textwidth]{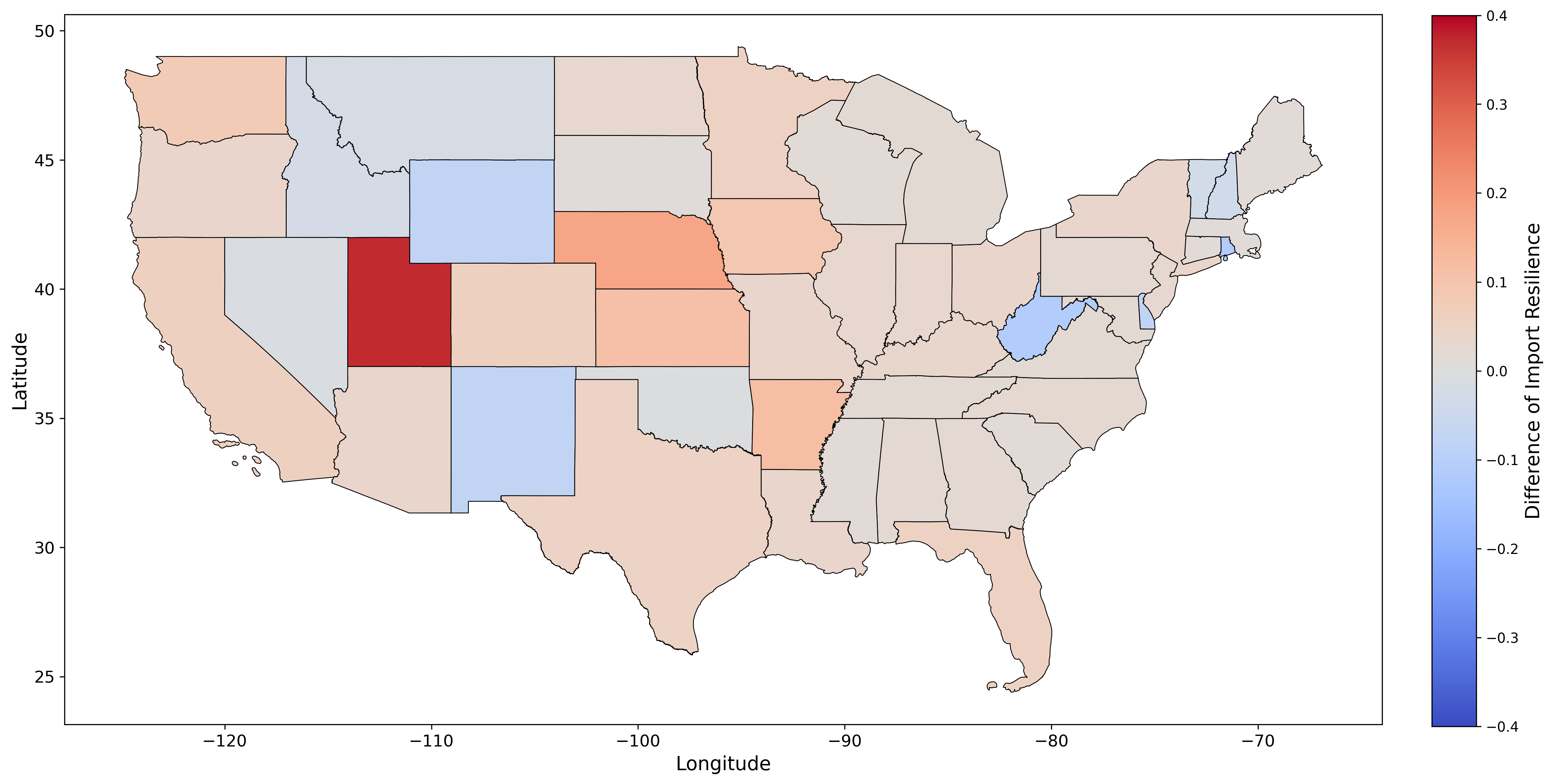}
	\end{minipage}
	\begin{minipage}[t]{0.48\textwidth}
		\centering
		\includegraphics[width=\textwidth]{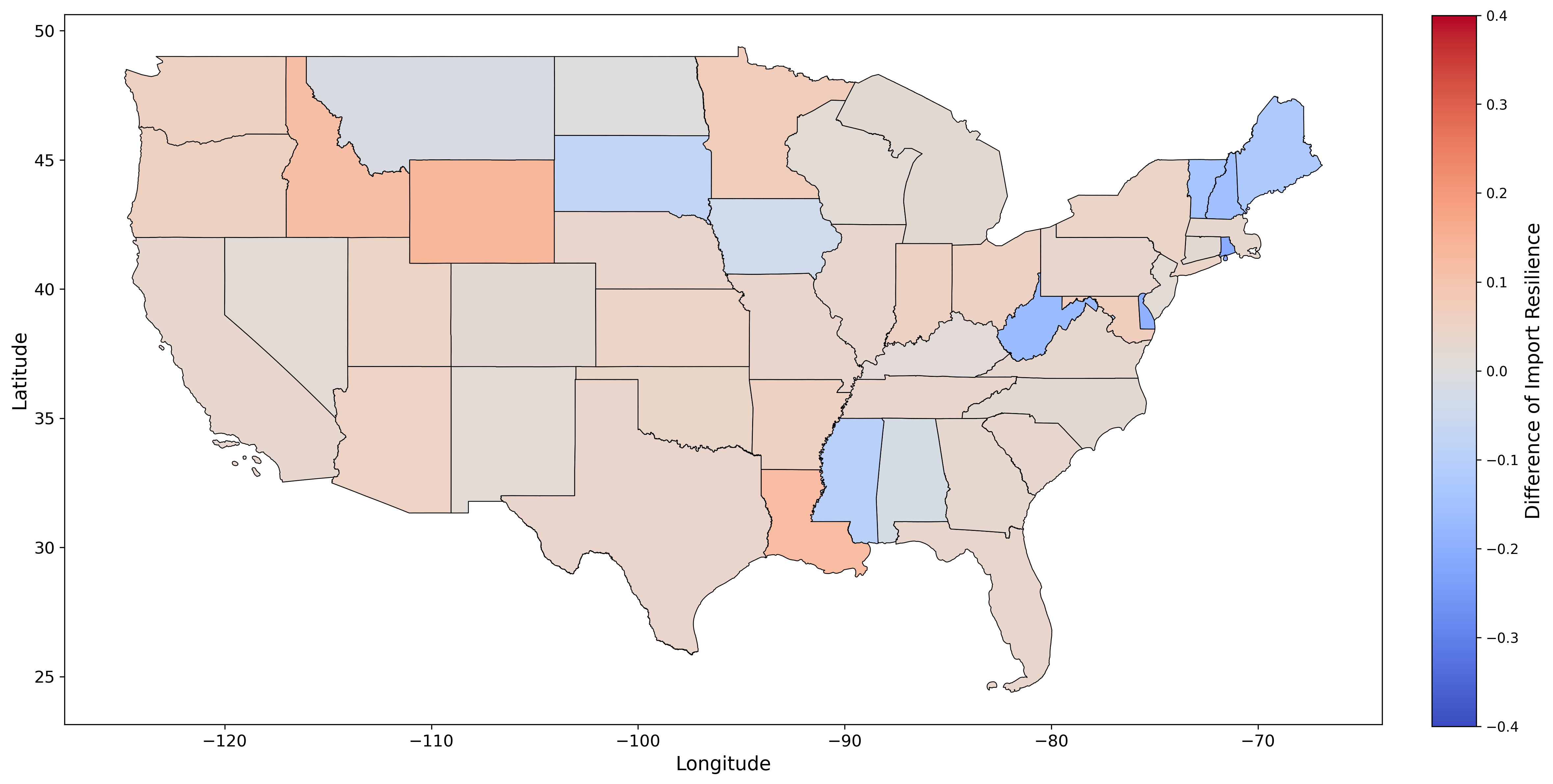}
	\end{minipage}
	\caption{The prediction error on 2012 (L) / 2017 (R) data by using centralized model trained on data generated by adjusting real 2012 data with noise = $0.3$}
	\label{fig:0.3_CT}
\end{figure*}

\begin{figure*}[!h]
	\centering
	\begin{minipage}[t]{0.48\textwidth}
		\centering
		\includegraphics[width=\textwidth]{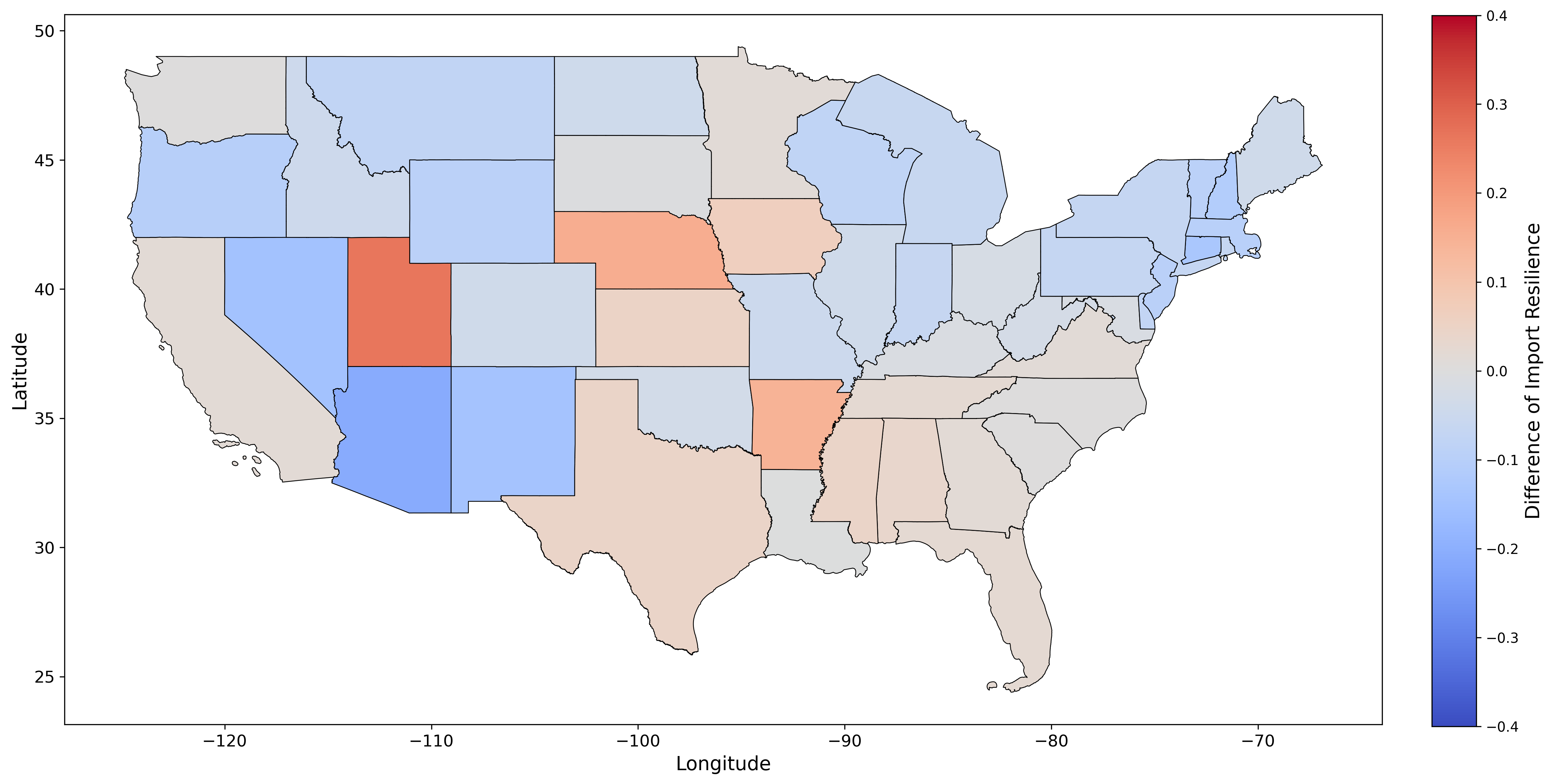}
	\end{minipage}
	\begin{minipage}[t]{0.48\textwidth}
		\centering
		\includegraphics[width=\textwidth]{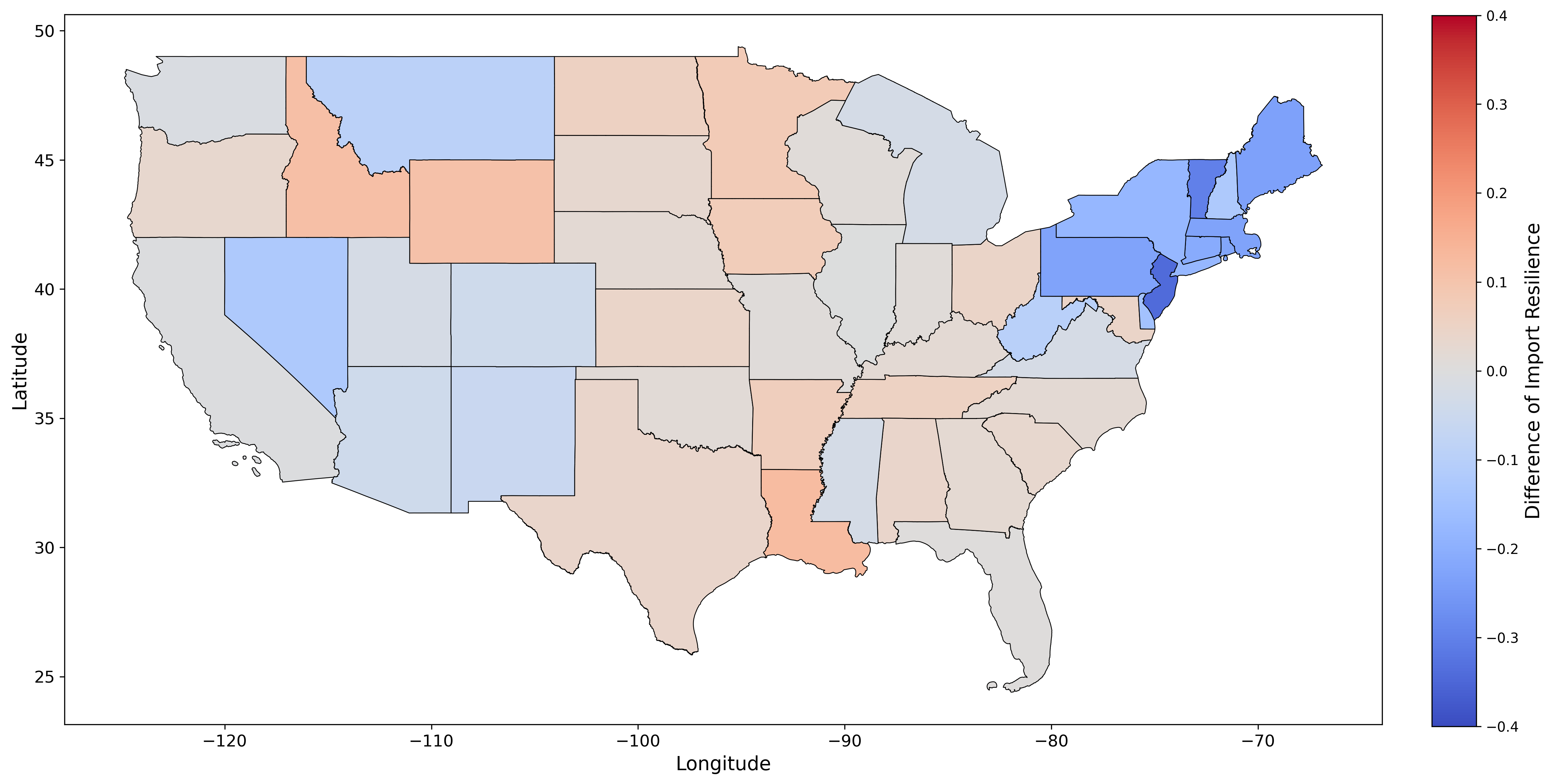}
	\end{minipage}
	\caption{The prediction error on 2012 (L) / 2017 (R) data by using federated learning model trained on silo data generated by adjusting real 2012 data noise = $0.3$}
	\label{fig:0.3_FL_silo}
\end{figure*}

In centralized training (Figure \ref{fig:0.3_CT}), the full graph is used both for training and evaluation. Consequently, there isn't any significant overestimation or underestimation observable from the graph. In the case of FLEE-GNN (Figure \ref{fig:0.3_FL_silo}), the silo sub-networks and ground truth label from the full graph are used for training. This simulates the scenario where we only have access to silo data in the real world due to privacy constraints. Yet, from that silo data, we still aim to recover the original true resilience of the full graph. During the training, our model tends to overestimate its resilience. However, this overestimation is offset by the lack of information in the actual global graph during evaluation. Through such a strategy, we also achieve balanced predictions when using FLEE-GNN without significant overestimation or underestimation.

\subsection{Absolute Prediction Error}

While the relative resilience difference captures the proportionality of errors, highlighting under- or over-estimations, we also examine the absolute prediction error, which quantifies the absolute deviations. For each model evaluated on data from both 2012 (Table \ref{2012_abs}) and 2017 (Table \ref{2017_abs}), we calculated the mean, standard deviation, minimum, 25th percentile, median (50th percentile), 75th percentile, and maximum values. The results indicate that both centralized training CT-GNN and federated training FLEE-GNN (with smaller prediction errors), trained on datasets generated with varying noise ratios, outperformed the entropy-based method across all metrics.

Another trend we observed from the results is that the performance tends to degrade slightly as we increase the noise ratio. This can be attributed to the fact that by injecting more noise, the generated food flow data becomes less representative of real data. Modern machine learning techniques often encounter this out-of-distribution issue \cite{yang2021generalized}. Furthermore, when comparing results between 2012 and 2017, the model maintains its performance even when evaluated on the 2017 food flow data. This is notable since the training data comes solely from the 2012 food flow dataset. Despite this, the model still performs admirably in predicting resilience for the food flow network in a different time period, demonstrating its versatility and effectiveness.

\begin{table*}[h]
	\scalebox{1.0}{
		\begin{tabular}{c|c|cccc|cccc|cccc}
			\textbf{Noise}   & \textbf{0}       & \multicolumn{4}{c|}{\textbf{0.1}}                                                                                    & \multicolumn{4}{c|}{\textbf{0.3}}                                                                                             & \multicolumn{4}{c}{\textbf{0.5}}                                                                                     \\ \hline
			\textbf{Dataset} & \textbf{SILO}    & \multicolumn{2}{c|}{\textbf{WHOLE}}                                 & \multicolumn{2}{c|}{\textbf{SILO}}             & \multicolumn{2}{c|}{\textbf{WHOLE}}                                 & \multicolumn{2}{c|}{\textbf{SILO}}             & \multicolumn{2}{c|}{\textbf{WHOLE}}                                 & \multicolumn{2}{c}{\textbf{SILO}}              \\ \hline
			\textbf{Method}  & \textbf{Entropy} & \multicolumn{1}{c|}{\textbf{CT}} & \multicolumn{1}{c|}{\textbf{FL}} & \multicolumn{1}{c|}{\textbf{CT}} & \textbf{FL} & \multicolumn{1}{c|}{\textbf{CT}} & \multicolumn{1}{c|}{\textbf{FL}} & \multicolumn{1}{c|}{\textbf{CT}} & \textbf{FL} & \multicolumn{1}{c|}{\textbf{CT}} & \multicolumn{1}{c|}{\textbf{FL}} & \multicolumn{1}{c|}{\textbf{CT}} & \textbf{FL} \\ \hline
			\textbf{mean}    & 0.1216           & \multicolumn{1}{c|}{0.0815}      & \multicolumn{1}{c|}{0.0628}      & \multicolumn{1}{c|}{0.2250}      & 0.0640      & \multicolumn{1}{c|}{0.0515}      & \multicolumn{1}{c|}{0.0685}      & \multicolumn{1}{c|}{0.1359}      & 0.0635      & \multicolumn{1}{c|}{0.0569}      & \multicolumn{1}{c|}{0.0746}      & \multicolumn{1}{c|}{0.1456}      & 0.0620      \\ \hline
			\textbf{std}     & 0.1475           & \multicolumn{1}{c|}{0.0504}      & \multicolumn{1}{c|}{0.0677}      & \multicolumn{1}{c|}{0.0745}      & 0.0548      & \multicolumn{1}{c|}{0.0575}      & \multicolumn{1}{c|}{0.0689}      & \multicolumn{1}{c|}{0.0688}      & 0.0556      & \multicolumn{1}{c|}{0.0615}      & \multicolumn{1}{c|}{0.0709}      & \multicolumn{1}{c|}{0.0746}      & 0.0580      \\ \hline
			\textbf{min}     & 0.0028           & \multicolumn{1}{c|}{0.0004}      & \multicolumn{1}{c|}{0.0004}      & \multicolumn{1}{c|}{0.0311}      & 0.0019      & \multicolumn{1}{c|}{0.0054}      & \multicolumn{1}{c|}{0.0036}      & \multicolumn{1}{c|}{0.0120}      & 0.0006      & \multicolumn{1}{c|}{0.0024}      & \multicolumn{1}{c|}{0.0117}      & \multicolumn{1}{c|}{0.0049}      & 0.0036      \\ \hline
			\textbf{25\%}    & 0.0246           & \multicolumn{1}{c|}{0.0366}      & \multicolumn{1}{c|}{0.0360}      & \multicolumn{1}{c|}{0.1733}      & 0.0249      & \multicolumn{1}{c|}{0.0205}      & \multicolumn{1}{c|}{0.0378}      & \multicolumn{1}{c|}{0.0874}      & 0.0209      & \multicolumn{1}{c|}{0.0236}      & \multicolumn{1}{c|}{0.0446}      & \multicolumn{1}{c|}{0.1002}      & 0.0204      \\ \hline
			\textbf{50\%}    & 0.0797           & \multicolumn{1}{c|}{0.0838}      & \multicolumn{1}{c|}{0.0458}      & \multicolumn{1}{c|}{0.2346}      & 0.0545      & \multicolumn{1}{c|}{0.0347}      & \multicolumn{1}{c|}{0.0533}      & \multicolumn{1}{c|}{0.1397}      & 0.0477      & \multicolumn{1}{c|}{0.0378}      & \multicolumn{1}{c|}{0.0564}      & \multicolumn{1}{c|}{0.1432}      & 0.0469      \\ \hline
			\textbf{75\%}    & 0.1574           & \multicolumn{1}{c|}{0.1100}      & \multicolumn{1}{c|}{0.0590}      & \multicolumn{1}{c|}{0.2627}      & 0.0785      & \multicolumn{1}{c|}{0.0596}      & \multicolumn{1}{c|}{0.0663}      & \multicolumn{1}{c|}{0.1916}      & 0.0921      & \multicolumn{1}{c|}{0.0682}      & \multicolumn{1}{c|}{0.0726}      & \multicolumn{1}{c|}{0.1993}      & 0.0935      \\ \hline
			\textbf{max}     & 0.6881           & \multicolumn{1}{c|}{0.2008}      & \multicolumn{1}{c|}{0.4482}      & \multicolumn{1}{c|}{0.3823}      & 0.2361      & \multicolumn{1}{c|}{0.3694}      & \multicolumn{1}{c|}{0.4612}      & \multicolumn{1}{c|}{0.3074}      & 0.2626      & \multicolumn{1}{c|}{0.3775}      & \multicolumn{1}{c|}{0.4769}      & \multicolumn{1}{c|}{0.3063}      & 0.3093     
		\end{tabular}
	}
	\caption{Absolute Prediction Error for 2012 Data (CT: Centralized Training; FL: Federating Learning)}
	\label{2012_abs}
\end{table*}

\begin{table*}[h]
	\scalebox{1.0}{
		\begin{tabular}{c|c|cccc|cccc|cccc}
			\textbf{Noise}   & \textbf{0} & \multicolumn{4}{c|}{\textbf{0.1}}                                                                                    & \multicolumn{4}{c|}{\textbf{0.3}}                                                                                         & \multicolumn{4}{c}{\textbf{0.5}}                                                                                              \\ \hline
			\textbf{Dataset} & \textbf{SILO}    & \multicolumn{2}{c|}{\textbf{WHOLE}}                                 & \multicolumn{2}{c|}{\textbf{SILO}}             & \multicolumn{2}{c|}{\textbf{WHOLE}}                                 & \multicolumn{2}{c|}{\textbf{SILO}}             & \multicolumn{2}{c|}{\textbf{WHOLE}}                                 & \multicolumn{2}{c}{\textbf{SILO}}              \\ \hline
			\textbf{Method}  & \textbf{Entropy}       & \multicolumn{1}{c|}{\textbf{CT}} & \multicolumn{1}{c|}{\textbf{FL}} & \multicolumn{1}{c|}{\textbf{CT}} & \textbf{FL} & \multicolumn{1}{c|}{\textbf{CT}} & \multicolumn{1}{c|}{\textbf{FL}} & \multicolumn{1}{c|}{\textbf{CT}} & \textbf{FL} & \multicolumn{1}{c|}{\textbf{CT}} & \multicolumn{1}{c|}{\textbf{FL}} & \multicolumn{1}{c|}{\textbf{CT}} & \textbf{FL} \\ \hline
			\textbf{mean}    & 0.1193           & \multicolumn{1}{c|}{0.0570}      & \multicolumn{1}{c|}{0.0737}      & \multicolumn{1}{c|}{0.1976}      & 0.0835      & \multicolumn{1}{c|}{0.0599}      & \multicolumn{1}{c|}{0.0751}      & \multicolumn{1}{c|}{0.1988}      & 0.0808      & \multicolumn{1}{c|}{0.0612}      & \multicolumn{1}{c|}{0.0749}      & \multicolumn{1}{c|}{0.2229}      & 0.0852      \\ \hline
			\textbf{std}     & 0.1480           & \multicolumn{1}{c|}{0.0500}      & \multicolumn{1}{c|}{0.0484}      & \multicolumn{1}{c|}{0.1097}      & 0.0925      & \multicolumn{1}{c|}{0.0496}      & \multicolumn{1}{c|}{0.0505}      & \multicolumn{1}{c|}{0.1120}      & 0.0836      & \multicolumn{1}{c|}{0.0556}      & \multicolumn{1}{c|}{0.0515}      & \multicolumn{1}{c|}{0.1108}      & 0.0787      \\ \hline
			\textbf{min}     & 0.0022           & \multicolumn{1}{c|}{0.0008}      & \multicolumn{1}{c|}{0.0108}      & \multicolumn{1}{c|}{0.0250}      & 0.0018      & \multicolumn{1}{c|}{0.0042}      & \multicolumn{1}{c|}{0.0308}      & \multicolumn{1}{c|}{0.0011}      & 0.0022      & \multicolumn{1}{c|}{0.0042}      & \multicolumn{1}{c|}{0.0297}      & \multicolumn{1}{c|}{0.0088}      & 0.0067      \\ \hline
			\textbf{25\%}    & 0.0305           & \multicolumn{1}{c|}{0.0251}      & \multicolumn{1}{c|}{0.0453}      & \multicolumn{1}{c|}{0.1305}      & 0.0227      & \multicolumn{1}{c|}{0.0290}      & \multicolumn{1}{c|}{0.0456}      & \multicolumn{1}{c|}{0.1163}      & 0.0228      & \multicolumn{1}{c|}{0.0266}      & \multicolumn{1}{c|}{0.0457}      & \multicolumn{1}{c|}{0.1327}      & 0.0284      \\ \hline
			\textbf{50\%}    & 0.0652           & \multicolumn{1}{c|}{0.0334}      & \multicolumn{1}{c|}{0.0628}      & \multicolumn{1}{c|}{0.1744}      & 0.0419      & \multicolumn{1}{c|}{0.0396}      & \multicolumn{1}{c|}{0.0612}      & \multicolumn{1}{c|}{0.1747}      & 0.0446      & \multicolumn{1}{c|}{0.0394}      & \multicolumn{1}{c|}{0.0612}      & \multicolumn{1}{c|}{0.2104}      & 0.0538      \\ \hline
			\textbf{75\%}    & 0.1292           & \multicolumn{1}{c|}{0.0621}      & \multicolumn{1}{c|}{0.0919}      & \multicolumn{1}{c|}{0.2601}      & 0.0975      & \multicolumn{1}{c|}{0.0665}      & \multicolumn{1}{c|}{0.0921}      & \multicolumn{1}{c|}{0.2760}      & 0.1139      & \multicolumn{1}{c|}{0.0642}      & \multicolumn{1}{c|}{0.0885}      & \multicolumn{1}{c|}{0.3200}      & 0.1093      \\ \hline
			\textbf{max}     & 0.7940           & \multicolumn{1}{c|}{0.1848}      & \multicolumn{1}{c|}{0.3058}      & \multicolumn{1}{c|}{0.5014}      & 0.3832      & \multicolumn{1}{c|}{0.2094}      & \multicolumn{1}{c|}{0.3309}      & \multicolumn{1}{c|}{0.4845}      & 0.3424      & \multicolumn{1}{c|}{0.2327}      & \multicolumn{1}{c|}{0.3390}      & \multicolumn{1}{c|}{0.4601}      & 0.3352     
		\end{tabular}
	}
	\caption{Absolute Prediction Error for 2017 Data (CT: Centralized Training; FL: Federating Learning)}
	\label{2017_abs}
\end{table*}

\subsection{Rank Evaluation}
Rather than focusing solely on the specific resilience provided by the model, we are also interested in how the rankings of node resilience in food supply networks are maintained. Specifically, we aim to determine whether a state with high or low resilience will consistently show high or low resilience values in comparison to other states in the FLEE-GNN model's output. To assess the consistency between predicted and actual state rankings based on their resilience values, we sorted the states accordingly and computed the following concordance metrics:
\begin{enumerate}
	\item The coincidence rate between the top n\% ranked nodes of actual data and the top n\% ranked nodes of predicted data.
	\item Pearson’s correlation coefficient.
	\item Spearman’s rank correlation coefficient.
\end{enumerate}

From the results (Table \ref{tab:rank_eval_2012} and Table \ref{tab:rank_eval_2017}), it is evident that 
the machine learning-based method also boasts a superior coincidence rate at different noise levels of data, signifying that these models offer higher consistency and alignment with the actual resilience values in the original dataset. Additionally, a higher Spearman's rho indicates that the rankings from the predictions of CT-GNN and FLEE-GNN closely match the actual rankings. Likewise, a larger Pearson's R indicates a robust linear relationship between predicted and actual resilience values. This means that as the actual resilience values rise (or fall), the model's predictions will also increase (or decrease) in a similar fashion. 

\begin{table*}[h]
	\scalebox{0.85}{
		\begin{tabular}{c|c|cccc|cccc|cccc}
			\textbf{Noise}                           & \textbf{0}       & \multicolumn{4}{c|}{\textbf{0.1}}                                                                                    & \multicolumn{4}{c|}{0.3}                                                                                             & \multicolumn{4}{c}{0.5}                                                                                              \\ \hline
			\textbf{Dataset}                         & \textbf{SILO}    & \multicolumn{2}{c|}{\textbf{WHOLE}}                                 & \multicolumn{2}{c|}{\textbf{SILO}}             & \multicolumn{2}{c|}{\textbf{WHOLE}}                                 & \multicolumn{2}{c|}{\textbf{SILO}}             & \multicolumn{2}{c|}{\textbf{WHOLE}}                                 & \multicolumn{2}{c}{\textbf{SILO}}              \\ \hline
			\textbf{Method}                          & \textbf{Entropy} & \multicolumn{1}{c|}{\textbf{CT}} & \multicolumn{1}{c|}{\textbf{FL}} & \multicolumn{1}{c|}{\textbf{CT}} & \textbf{FL} & \multicolumn{1}{c|}{\textbf{CT}} & \multicolumn{1}{c|}{\textbf{FL}} & \multicolumn{1}{c|}{\textbf{CT}} & \textbf{FL} & \multicolumn{1}{c|}{\textbf{CT}} & \multicolumn{1}{c|}{\textbf{FL}} & \multicolumn{1}{c|}{\textbf{CT}} & \textbf{FL} \\ \hline
			\textbf{Coincidence / Recall (Top 10\%)} & 0.0000           & \multicolumn{1}{c|}{0.5000}      & \multicolumn{1}{c|}{0.3333}      & \multicolumn{1}{c|}{0.1667}      & 0.5000      & \multicolumn{1}{c|}{0.5000}      & \multicolumn{1}{c|}{0.5000}      & \multicolumn{1}{c|}{0.3333}      & 0.5000      & \multicolumn{1}{c|}{0.5000}      & \multicolumn{1}{c|}{0.5000}      & \multicolumn{1}{c|}{0.3333}      & 0.3333      \\ \hline
			\textbf{Coincidence / Recall (Top 30\%)} & 0.4375           & \multicolumn{1}{c|}{0.5625}      & \multicolumn{1}{c|}{0.7500}      & \multicolumn{1}{c|}{0.5625}      & 0.5000      & \multicolumn{1}{c|}{0.7500}      & \multicolumn{1}{c|}{0.7500}      & \multicolumn{1}{c|}{0.5625}      & 0.4375      & \multicolumn{1}{c|}{0.7500}      & \multicolumn{1}{c|}{0.7500}      & \multicolumn{1}{c|}{0.5625}      & 0.4375      \\ \hline
			\textbf{Coincidence / Recall (Top 50\%)} & 0.6154           & \multicolumn{1}{c|}{0.8462}      & \multicolumn{1}{c|}{0.8462}      & \multicolumn{1}{c|}{0.7692}      & 0.8077      & \multicolumn{1}{c|}{0.8462}      & \multicolumn{1}{c|}{0.8462}      & \multicolumn{1}{c|}{0.7308}      & 0.7308      & \multicolumn{1}{c|}{0.8077}      & \multicolumn{1}{c|}{0.7692}      & \multicolumn{1}{c|}{0.7308}      & 0.6923      \\ \hline
			\textbf{Pearson R}                       & 0.4805           & \multicolumn{1}{c|}{0.8098}      & \multicolumn{1}{c|}{0.7066}      & \multicolumn{1}{c|}{0.5251}      & 0.7093      & \multicolumn{1}{c|}{0.7272}      & \multicolumn{1}{c|}{0.6979}      & \multicolumn{1}{c|}{0.5680}      & 0.6619      & \multicolumn{1}{c|}{0.7068}      & \multicolumn{1}{c|}{0.6673}      & \multicolumn{1}{c|}{0.5620}      & 0.6380      \\ \hline
			\textbf{Spearman rho}                      & 0.3135           & \multicolumn{1}{c|}{0.8110}      & \multicolumn{1}{c|}{0.8642}      & \multicolumn{1}{c|}{0.5738}      & 0.6624      & \multicolumn{1}{c|}{0.8251}      & \multicolumn{1}{c|}{0.8338}      & \multicolumn{1}{c|}{0.5932}      & 0.6118      & \multicolumn{1}{c|}{0.8195}      & \multicolumn{1}{c|}{0.8121}      & \multicolumn{1}{c|}{0.5600}      & 0.5542     
		\end{tabular}
	}
	\caption{Rank Evaluation Metrics for 2012 CFS Data (CT: Centralized Training; FL: Federating Learning)}
	\label{tab:rank_eval_2012}
\end{table*}

\begin{table*}[h]
	\scalebox{0.85}{
		\begin{tabular}{c|c|cccc|cccc|cccc}
			\textbf{Noise}                           & \textbf{0}       & \multicolumn{4}{c|}{\textbf{0.1}}                                                                                    & \multicolumn{4}{c|}{\textbf{0.3}}                                                                                    & \multicolumn{4}{c}{\textbf{0.5}}                                                                                     \\ \hline
			\textbf{Dataset}                         & \textbf{SILO}    & \multicolumn{2}{c|}{\textbf{WHOLE}}                                 & \multicolumn{2}{c|}{\textbf{SILO}}             & \multicolumn{2}{c|}{\textbf{WHOLE}}                                 & \multicolumn{2}{c|}{\textbf{SILO}}             & \multicolumn{2}{c|}{\textbf{WHOLE}}                                 & \multicolumn{2}{c}{\textbf{SILO}}              \\ \hline
			\textbf{Method}                          & \textbf{Entropy} & \multicolumn{1}{c|}{\textbf{CT}} & \multicolumn{1}{c|}{\textbf{FL}} & \multicolumn{1}{c|}{\textbf{CT}} & \textbf{FL} & \multicolumn{1}{c|}{\textbf{CT}} & \multicolumn{1}{c|}{\textbf{FL}} & \multicolumn{1}{c|}{\textbf{CT}} & \textbf{FL} & \multicolumn{1}{c|}{\textbf{CT}} & \multicolumn{1}{c|}{\textbf{FL}} & \multicolumn{1}{c|}{\textbf{CT}} & \textbf{FL} \\ \hline
			\textbf{Coincidence / Recall (Top 10\%)} & 0.2000           & \multicolumn{1}{c|}{0.4000}      & \multicolumn{1}{c|}{0.4000}      & \multicolumn{1}{c|}{0.1667}      & 0.5000      & \multicolumn{1}{c|}{0.6000}      & \multicolumn{1}{c|}{0.4000}      & \multicolumn{1}{c|}{0.3333}      & 0.5000      & \multicolumn{1}{c|}{0.4000}      & \multicolumn{1}{c|}{0.4000}      & \multicolumn{1}{c|}{0.3333}      & 0.3333      \\ \hline
			\textbf{Coincidence / Recall (Top 30\%)} & 0.5333           & \multicolumn{1}{c|}{0.5333}      & \multicolumn{1}{c|}{0.6667}      & \multicolumn{1}{c|}{0.5625}      & 0.5000      & \multicolumn{1}{c|}{0.5333}      & \multicolumn{1}{c|}{0.6667}      & \multicolumn{1}{c|}{0.5625}      & 0.4375      & \multicolumn{1}{c|}{0.6667}      & \multicolumn{1}{c|}{0.6667}      & \multicolumn{1}{c|}{0.5625}      & 0.4375      \\ \hline
			\textbf{Coincidence / Recall (Top 50\%)} & 0.6000           & \multicolumn{1}{c|}{0.8800}      & \multicolumn{1}{c|}{0.8000}      & \multicolumn{1}{c|}{0.7692}      & 0.8077      & \multicolumn{1}{c|}{0.8800}      & \multicolumn{1}{c|}{0.8400}      & \multicolumn{1}{c|}{0.7308}      & 0.7308      & \multicolumn{1}{c|}{0.8400}      & \multicolumn{1}{c|}{0.8400}      & \multicolumn{1}{c|}{0.7308}      & 0.6923      \\ \hline
			\textbf{Pearson R}                       & 0.7704           & \multicolumn{1}{c|}{0.7710}      & \multicolumn{1}{c|}{0.7670}      & \multicolumn{1}{c|}{0.5251}      & 0.7093      & \multicolumn{1}{c|}{0.7886}      & \multicolumn{1}{c|}{0.7641}      & \multicolumn{1}{c|}{0.5680}      & 0.6619      & \multicolumn{1}{c|}{0.7811}      & \multicolumn{1}{c|}{0.7581}      & \multicolumn{1}{c|}{0.5620}      & 0.6380      \\ \hline
			\textbf{Spearman rho}                      & 0.4736           & \multicolumn{1}{c|}{0.8008}      & \multicolumn{1}{c|}{0.8210}      & \multicolumn{1}{c|}{0.5738}      & 0.6624      & \multicolumn{1}{c|}{0.8215}      & \multicolumn{1}{c|}{0.8317}      & \multicolumn{1}{c|}{0.5932}      & 0.6118      & \multicolumn{1}{c|}{0.8372}      & \multicolumn{1}{c|}{0.8206}      & \multicolumn{1}{c|}{0.5600}      & 0.5542     
		\end{tabular}
	}
	\caption{Rank Evaluation Metrics for 2017 CFS Data (CT: Centralized Training; FL: Federating Learning)}
	\label{tab:rank_eval_2017}
\end{table*}

\subsection{Robustness with Missing Data}

\begin{table*}[h]
	\scalebox{0.85}{
		\begin{tabular}{c|cccccccccccccccc}
			\textbf{Noise}   & \multicolumn{16}{c}{\textbf{0.3}}                                                                                                   \\ \hline
			\textbf{Columns} & \multicolumn{2}{c|}{\textbf{VAT}}                                   & \multicolumn{2}{c|}{\textbf{VT}}                                    & \multicolumn{2}{c|}{\textbf{VA}}                                    & \multicolumn{2}{c|}{\textbf{TA}}                                    & \multicolumn{2}{c|}{\textbf{V}}                                     & \multicolumn{2}{c|}{\textbf{T}}                                     & \multicolumn{2}{c|}{\textbf{A}}   & \multicolumn{2}{c}{\textbf{NONE}}              \\ \hline
			\textbf{Method}  & \multicolumn{1}{c|}{\textbf{CT}} & \multicolumn{1}{c|}{\textbf{FL}} & \multicolumn{1}{c|}{\textbf{CT}} & \multicolumn{1}{c|}{\textbf{FL}} & \multicolumn{1}{c|}{\textbf{CT}} & \multicolumn{1}{c|}{\textbf{FL}} & \multicolumn{1}{c|}{\textbf{CT}} & \multicolumn{1}{c|}{\textbf{FL}} & \multicolumn{1}{c|}{\textbf{CT}} & \multicolumn{1}{c|}{\textbf{FL}} & \multicolumn{1}{c|}{\textbf{CT}} & \multicolumn{1}{c|}{\textbf{FL}} & \multicolumn{1}{c|}{\textbf{CT}} & \multicolumn{1}{c|}{\textbf{FL}} & \multicolumn{1}{c|}{\textbf{CT}} & \textbf{FL} \\ \hline
			\textbf{mean}    & \multicolumn{1}{c|}{0.0599}      & \multicolumn{1}{c|}{0.0808}      & \multicolumn{1}{c|}{0.0600}      & \multicolumn{1}{c|}{0.0840}      & \multicolumn{1}{c|}{0.0574}      & \multicolumn{1}{c|}{0.0786}      & \multicolumn{1}{c|}{0.0602}      & \multicolumn{1}{c|}{0.0794}      & \multicolumn{1}{c|}{0.0601}      & \multicolumn{1}{c|}{0.0834}      & \multicolumn{1}{c|}{0.0792}      & \multicolumn{1}{c|}{0.0806}      & \multicolumn{1}{c|}{0.0583}      & \multicolumn{1}{c|}{0.0872}      & \multicolumn{1}{c|}{0.0793}      & 0.0978      \\ \hline
			\textbf{std}     & \multicolumn{1}{c|}{0.0496}      & \multicolumn{1}{c|}{0.0836}      & \multicolumn{1}{c|}{0.0549}      & \multicolumn{1}{c|}{0.0910}      & \multicolumn{1}{c|}{0.0540}      & \multicolumn{1}{c|}{0.0823}      & \multicolumn{1}{c|}{0.0648}      & \multicolumn{1}{c|}{0.0829}      & \multicolumn{1}{c|}{0.0595}      & \multicolumn{1}{c|}{0.0935}      & \multicolumn{1}{c|}{0.0864}      & \multicolumn{1}{c|}{0.0873}      & \multicolumn{1}{c|}{0.0672}      & \multicolumn{1}{c|}{0.1011}      & \multicolumn{1}{c|}{0.0864}      & 0.0991      \\ \hline
			\textbf{min}     & \multicolumn{1}{c|}{0.0042}      & \multicolumn{1}{c|}{0.0023}      & \multicolumn{1}{c|}{0.0053}      & \multicolumn{1}{c|}{0.0001}      & \multicolumn{1}{c|}{0.0004}      & \multicolumn{1}{c|}{0.0021}      & \multicolumn{1}{c|}{0.0013}      & \multicolumn{1}{c|}{0.0042}      & \multicolumn{1}{c|}{0.0011}      & \multicolumn{1}{c|}{0.0004}      & \multicolumn{1}{c|}{0.0006}      & \multicolumn{1}{c|}{0.0019}      & \multicolumn{1}{c|}{0.0011}      & \multicolumn{1}{c|}{0.0004}      & \multicolumn{1}{c|}{0.0001}      & 0.0001      \\ \hline
			\textbf{25\%}    & \multicolumn{1}{c|}{0.0290}      & \multicolumn{1}{c|}{0.0228}      & \multicolumn{1}{c|}{0.0257}      & \multicolumn{1}{c|}{0.0188}      & \multicolumn{1}{c|}{0.0227}      & \multicolumn{1}{c|}{0.0218}      & \multicolumn{1}{c|}{0.0162}      & \multicolumn{1}{c|}{0.0233}      & \multicolumn{1}{c|}{0.0226}      & \multicolumn{1}{c|}{0.0169}      & \multicolumn{1}{c|}{0.0156}      & \multicolumn{1}{c|}{0.0160}      & \multicolumn{1}{c|}{0.0142}      & \multicolumn{1}{c|}{0.0194}      & \multicolumn{1}{c|}{0.0194}      & 0.0236      \\ \hline
			\textbf{50\%}    & \multicolumn{1}{c|}{0.0396}      & \multicolumn{1}{c|}{0.0446}      & \multicolumn{1}{c|}{0.0402}      & \multicolumn{1}{c|}{0.0485}      & \multicolumn{1}{c|}{0.0371}      & \multicolumn{1}{c|}{0.0418}      & \multicolumn{1}{c|}{0.0289}      & \multicolumn{1}{c|}{0.0394}      & \multicolumn{1}{c|}{0.0403}      & \multicolumn{1}{c|}{0.0398}      & \multicolumn{1}{c|}{0.0321}      & \multicolumn{1}{c|}{0.0439}      & \multicolumn{1}{c|}{0.0259}      & \multicolumn{1}{c|}{0.0451}      & \multicolumn{1}{c|}{0.0421}      & 0.0557      \\ \hline
			\textbf{75\%}    & \multicolumn{1}{c|}{0.0665}      & \multicolumn{1}{c|}{0.1139}      & \multicolumn{1}{c|}{0.0625}      & \multicolumn{1}{c|}{0.1130}      & \multicolumn{1}{c|}{0.0743}      & \multicolumn{1}{c|}{0.1090}      & \multicolumn{1}{c|}{0.0913}      & \multicolumn{1}{c|}{0.1160}      & \multicolumn{1}{c|}{0.0611}      & \multicolumn{1}{c|}{0.1239}      & \multicolumn{1}{c|}{0.1287}      & \multicolumn{1}{c|}{0.1055}      & \multicolumn{1}{c|}{0.0760}      & \multicolumn{1}{c|}{0.1118}      & \multicolumn{1}{c|}{0.1314}      & 0.1335      \\ \hline
			\textbf{max}     & \multicolumn{1}{c|}{0.2094}      & \multicolumn{1}{c|}{0.3424}      & \multicolumn{1}{c|}{0.2344}      & \multicolumn{1}{c|}{0.3655}      & \multicolumn{1}{c|}{0.2202}      & \multicolumn{1}{c|}{0.3332}      & \multicolumn{1}{c|}{0.2421}      & \multicolumn{1}{c|}{0.3289}      & \multicolumn{1}{c|}{0.2452}      & \multicolumn{1}{c|}{0.3677}      & \multicolumn{1}{c|}{0.2938}      & \multicolumn{1}{c|}{0.3413}      & \multicolumn{1}{c|}{0.2571}      & \multicolumn{1}{c|}{0.3685}      & \multicolumn{1}{c|}{0.3131}      & 0.3724     
		\end{tabular}
	}
	\caption{Absolute Prediction Error of the ablation study for 2017 CFS Data (CT: Centralized Training; FL: Federating Learning)}
	\label{tab:ape_2017}
\end{table*}

Not only are we interested in the performance of FLEE-GNN at different time periods (2012/2017) and in different spaces (global transportation, silo transportation), but we are also keen on its performance across different missing data scenarios. In these experiments, we compare the absolute prediction error for CT-GNN trained with data generated from the 2012 food flow network and evaluated on the 2017 global network data, with that of FLEE-GNN trained with the same data and evaluated on 2017  sub-network silo data. However, during training, we consider different data combinations to simulate scenarios where certain edge features missing from a dataset.
\begin{itemize}
	\item V: Represents that the value of commodity food flow exists in edge features.
	\item A: Represents that the average transportation mile of commodity food flow exists in edge features.
	\item T: Represents that the tonnage mile of commodity food flow exists in edge features.
\end{itemize}

For example, "AT" means the graph structure of the food supply network remains the same, but instead of having access to all the shipment value, average transportation mile, and tonnage, we lose the shipment value information.

As seen in Table \ref{tab:ape_2017}, interestingly, both CT-GNN and FLEE-GNN maintain stability even with some missing  edge features. We do notice that some performance decreases when the data constrain the type of information the model can access. The maximum difference in terms of mean absolute prediction error for CT-GNN is $0.0793-0.0599=0.0195$ and for FLEE-GNN is $0.0978-0.0808=0.017$.

To some extent, this research proves that the most relevant information supporting current resilience measurement is embedded in the graph structure. In the extreme \textit{NONE} case (in Table \ref{tab:ape_2017}), all we retain is the basic graph structure (topological relations between nodes) without any edge features. Still, both CT-GNN and FLEE-GNN provide a satisfactory estimation of the resilience value. However, the original entropy-based method, which relies heavily on the edge attribution, fails to generate a meaningful resilience score in this case.

\subsection{Summary}
The silo-data-based resilience metrics using the entropy method revealed its constraints, particularly in its limited access to comprehensive linkages among geographical regions. While both the CT-GNN and FLEE-GNN demonstrated performance robustness against noise, and FLEE-GNN outperformed consistently, especially in higher noise scenarios. The Spearman rank correlation of the FLEE-GNN indicates a strong agreement with the ground-truth data, emphasizing the model's utility in accurately ranking the resilience of U.S. states in food supply. Given that the eventual aim is to bolster food security, a model like FLEE-GNN offers valuable insights into spatial network vulnerabilities and opportunities to improve supply network resilience. This could guide food systems policy-making, resource allocation and negotiation in critical scenarios in advance of cascading system failures. 

In summary, FLEE-GNN not only holds promise as an innovative tool for assessing food network resilience but also sets the stage for leveraging federated learning in a myriad of complex, geospatial AI applications. Future directions might delve deeper into optimizing FLEE-GNN and expanding its applicability to other domains.

\section{Future Work}

\subsection{Unsupervised Learning}

Our initial work utilized labels generated by the entropy-based method as the ground truth. However, this limits us to the upper bound provided by the entropy-based method. Unsupervised learning has demonstrated greate success in the field of image classification \cite{schmarje2021survey}, feature selection \cite{khanum2015survey}, association rule detection \cite{reshef2011detecting}, etc. Developing unsupervised learning techniques may allow us to exceed this bound and improve our system's predictive power.

\subsection{GeoKG Embedding and Semantic Extraction}

We currently input different attributes into the FLEE-GNN as scalar values. The model thus lacks an understanding of the differences between values, tons, average miles, temperature control, and other attributes in food supply transport. Moreover, it cannot extract comprehensive relationships between these attributes. To improve our model's predictive ability and interpretability, we need to devise a better explainable AI method for embedding GeoKG information directly into our system.

\subsection{Other Domains}
Moreover, the emphasis on federated learning within FLEE-GNN opens doors for its application in scenarios where data privacy and decentralization are paramount. This consideration will undoubtedly grow more significant in an era marked by heightened data protection regulations and the need for more inclusive, globally-spread research collaborations.

In conclusion, while the immediate implications of our work pertain to food supply networks, the FLEE-GNN framework offers a versatile blueprint with applications across a multitude of sectors. We eagerly anticipate the breakthroughs and innovations that can arise as we venture into other types of spatial networks and application domains.

\section{Conclusion}

In the face of increasing global food insecurity, understanding and fortifying the resilience of food supply networks is of paramount importance. Our study introduced the FLEE-GNN, a pioneering Federated Learning System for Edge-enhanced Graph Neural Network, adeptly tailored to navigate the complexities of spatial networks. Our comprehensive experimental design and evaluations elucidated the method's superiority over traditional entropy-based systems, especially in terms of robustness, scalability, and data privacy considerations. The novel integration of graph neural networks with the decentralized architecture of federated learning showcased how data privacy and model robustness can be harmoniously balanced when dealing with spatially heterogeneous sub-network datasets. The FLEE-GNN's performance, particularly in scenarios with noise, demonstrates its potential as a valuable tool for stakeholders across the global food supply network. Not only does it offer insights into the current state of the food system, but its predictive prowess can be instrumental in proactive decision-making and vulnerability mitigation.

Furthermore, the limitations of entropy methods, particularly with constrained data access, were evident. This highlights the pressing need for innovative, decentralized models like FLEE-GNN in our evolving digital age, where data is often fragmented and siloed. As we contemplate a future with increasing challenges to food security, AI tools like the FLEE-GNN will be indispensable. They promise a more informed, agile, and resilient approach to safeguarding our most crucial supply networks. While this study illuminates a promising path forward, it also beckons further exploration and refinement in the realm of federated learning and neural network integration for complex network analysis especially for public sector data. We are optimistic that this fusion of geospatial AI technologies can serve as a linchpin in the global endeavor to ensure food security for all.

\section*{Data and code availability statement}
The data and codes that support the findings of this study are available at the following link on GitHub: \url{https://github.com/GeoDS/FLEE-GNN/}.

\begin{acks}
We acknowledge the funding support from the National Science Foundation funded AI institute [Grant No. 2112606] for Intelligent Cyberinfrastructure with Computational Learning in the Environment (ICICLE). Any opinions, findings, and conclusions or recommendations expressed in this material are those of the author(s) and do not necessarily reflect the views of the funder(s). 
\end{acks}

\bibliographystyle{ACM-Reference-Format}
\bibliography{references}




\end{document}